\documentclass[runningheads]{llncs}
\usepackage{graphicx}

\usepackage{hyperref}
\usepackage{amssymb}
\usepackage{multirow}
\usepackage{booktabs}
\usepackage{subfigure}
\usepackage{bbding}

\begin{document}
	\title{Kernel Attention Transformer (KAT) for Histopathology Whole Slide Image Classification}
	\titlerunning{Kernel attention Transformer (KAT) for histopathology WSI classification}
	%
	\author{Yushan Zheng \inst{1(}\Envelope\inst{)} \and Jun Li \inst{2} \and Jun Shi \inst{3} \and Fengying Xie \inst{2}\and Zhiguo Jiang \inst{2}}
	\authorrunning{Y. Zheng et al.}
	\institute{School of Engineering Medicine, Beijing Advanced Innovation Center on Biomedical Engineering, Beihang University, Beijing 100191, China.\\\email{yszheng@buaa.edu.cn} \and Image Processing Center, School of Astronautics, Beihang University, Beijing, 102206, China. \and
		School of Software, Hefei University of Technology, Hefei 230601, China.}

	\maketitle              
	\begin{abstract}
		Transformer has been widely used in histopathology whole slide image (WSI) classification for the purpose of tumor grading, prognosis analysis, etc. However, the design of token-wise self-attention and positional embedding strategy in the common Transformer limits the effectiveness and efficiency in the application to gigapixel histopathology images. In this paper, we propose a kernel attention Transformer (KAT) for histopathology WSI classification. The information transmission of the tokens is achieved by cross-attention between the tokens and a set of kernels related to a set of positional anchors on the WSI. Compared to the common Transformer structure, the proposed KAT can better describe the hierarchical context information of the local regions of the WSI and meanwhile maintains a lower computational complexity. The proposed method was evaluated on a gastric dataset with 2040 WSIs and an endometrial dataset with 2560 WSIs, and was compared with 6 state-of-the-art methods. The experimental results have demonstrated the proposed KAT is effective and efficient in the task of histopathology WSI classification and is superior to the state-of-the-art methods. The code is available at \url{https://github.com/zhengyushan/kat}.
		
	\end{abstract}
	\section{Introduction}
	Histopathology whole slide image (WSI) classification based on image processing and deep learning has proven effective to building computer-aided applications for cancer screening \cite{song2020clinically,yu2021large}, tumor grading \cite{bulten2022artificial}, prognosis analysis \cite{fu2020pan}, gene mutant prediction \cite{yamashita2021deep}, etc.
	
	Recently, Vision Transformer (ViT) \cite{dosovitskiy2020image}, the extensively studied model in natural scene image recognition, was introduced to this problem. The WSI classification can be achieved by regarding the WSI local features as tokens. Theoretically, the self-attention mechanism of Transformer enables it to detect the useful relations of the local features for WSI recognition. The recent studies \cite{gao2021instance,chen2021whole,li2021dt} have proven Transformer-based models can further improve the WSI classification accuracy when compared to the previous methods based on convolution neural network (CNN) \cite{tellez2019neural}, recurrent neural network (RNN) \cite{campanella2019clinical}, and multiple instance learning (MIL) \cite{lu2021data}. Nevertheless, the calculation flowchart of self-attention, the main operation in Transformer, occurs notable problem when applied for the WSI containing gigapixels. 
	Firstly, the positional embedding strategy of ViT, which is designed for images in fixed size and shape, cannot consistently describe the structural information of WSI. Secondly, the self-attention operation permits equivalent conjunction of tokens in every stage of transformer. The equivalent conjunction usually causes over-smoothing in the token representations and thereby does harm the learning of the local patterns of the WSI. Furthermore, the computational complexity of the self-attention operation is $\mathcal{O}(n^2)$ to the number of tokens $n$. The inference of the Transformer becomes rather inefficient when facing a WSI with thousands of tokens. These problems have affected the efficiency and accuracy of the Transformer-based method for WSI classification.
	
	\begin{figure}[t]
		\includegraphics[width=\textwidth]{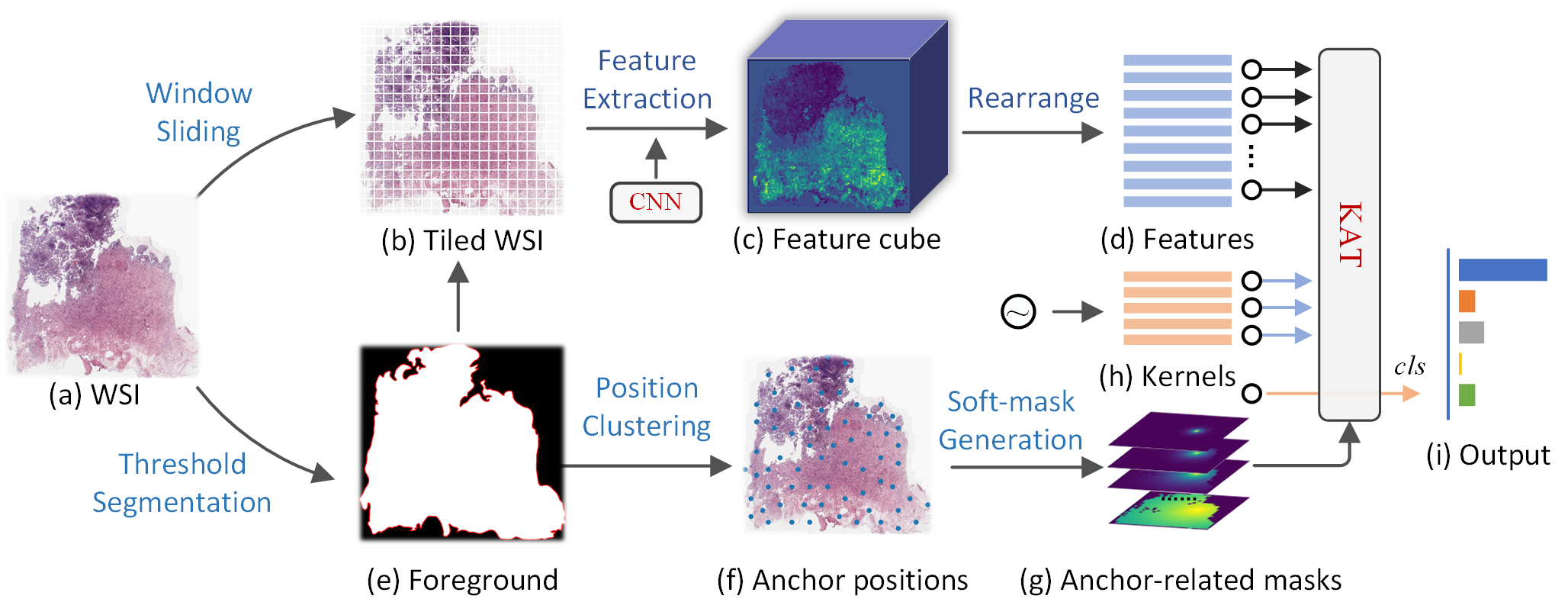}
		\caption{The proposed  framework for WSI classification, where (a-d) is the process of feature extraction for the WSI, (a-g) illustrates the flowchart of anchor position detection and mask generation, (h) is a set of trainable kernel vectors that related to the anchor positions, and KAT is the proposed kernel attention network that is detailed in seciton \ref{S.kat} and Fig. \ref{F.network}.} \label{F.framework}
	\end{figure}
	In this paper, we propose a novel Transformer model named kernel attention Transformer (KAT) and an integrated framework for histopathology whole slide image classification, which is illustrated in Fig. \ref{F.framework}. Compared to the common Transformer structure, the proposed KAT can describe hierarchical context information of the local regions of the WSI and thereby is more effective for histopathology WSI analysis. Meanwhile, the kernel-based cross-attention paradigm maintains a near-linear computational complexity to the size of the WSI. The proposed method was evaluated on a gastric dataset with 2040 WSIs and an endometrial dataset with 2560 WSIs, and was compared with 6 state-of-the-art methods \cite{chen2021whole,lu2021data,shao2021transmil,xiong2021nystromformer,dosovitskiy2020image,zheng2022encoding}. The experimental results have demonstrated the proposed KAT is effective and efficient in the task of histopathology WSI classification and is superior to the state-of-the-art methods.
	
	The contribution of the paper can be summarized in two aspects.
	\textbf{(1)} A novel Transformer-based structure named kernel attention Transformer (KAT) is proposed. Unlike the common Transformer, the information transmission of the tokens is achieved by cross-attention between the tokens and a set of kernels that are bounded to a set of positional anchors on the WSI. The experimental results show that the kernel-based cross-attention in KAT contributes to a competitive performance for WSI classification. Furthermore, it significantly relieves the burden of the computation device in both the training and the application stages. Furthermore, 
	\textbf{(2)} We design a flowchart to generate hierarchical positional masks to define multi-scale WSI regions. The positional masks are calculated based on the spatial allocations of the tokens and then bound with the kernels in KAT. It makes the KAT able to learn hierarchical representations from the local to the global scale of the WSI and thereby delivers better WSI classification performance.
	
	\section{Method}
	
	\subsection{Pre-processing and Data preparation}
	The blank regions of the WSI are removed beforehand by a threshold on the intensity, for these regions are less informative for diagnosis. Then, a tissue mask for the WSI is generated by filling the cavities of the foreground of the threshold output (as shown in Fig. \ref{F.framework}e). Based on the mask, the tissue region is divided into patches following the non-overlapping sliding window paradigm (Fig. \ref{F.framework}b). Next, a CNN is trained to extract the patch features. The structure of CNN was EfficientNet-b0 \cite{tan2019efficientnet} because of its lightweight and comprehensive performance. To meet the consensus of annotation-free modeling in the domain of histopathology WSI analysis, the EfficientNet-b0 is trained by contrastive representation learning proposed in BYOL \cite{grill2020bootstrap}. 

	One motivation for the kernel-based scheme is to refine the local information communication of the Transformer. Therefore, we designed a flowchart to generate hierarchical masks to guide information transmission in different stages. The features within the foreground of the tissue mask were rearranged to a feature matrix (as shown in Fig. \ref{F.framework}d) which is formulated as $\mathbf{X}\in\mathbb{R}^{n_p\times d_f}$, where $n_p$ denotes the number of tissue-related patches on the WSI, $\mathbf{x}_i\in\mathbb{R}^{d_f}$ is the $i$-th row of the $\mathbf{X}$ that represents the feature of the $i$-th tissue-related patch. Correspondingly, we define $p(\mathbf{x}_i)=(m_i, n_i)^\mathrm{T}$ to represent the patch-wise coordinate of the feature $\mathbf{x}_i$ on the WSI. Then, we extracted a set of anchors based on the spatially clustering property of the features. Specifically, the K-means algorithm is applied to clustering $\{p(\mathbf{x}_i)| i=1,2,...,n_p\}$ into $K$ centers. Then, the most nearest position to each center is recognized as an anchor position, which is represented as $\mathbf{c}_k=(m_k, n_k)^\mathrm{T},k=1,2...,K$, as shown in Fig. \ref{F.framework}f.
	
	Afterward, we define weighting masks for each anchor position based on the spatial distance between features and the anchor. Specifically, the weight of the $k$-th anchor and the $i$-th feature is calculated by the equation
	\begin{equation}
	m_{ki}(\delta)= \exp(-\|p(\mathbf{x}_i)-\mathbf{c}_k\|^2_2/2\delta^2),
	\end{equation}
	where $\delta$ controls the scale of the Gaussian-like mask. By computing the weights between all the features and anchors, we obtain the mask matrix $\mathbf{M}(\delta)\in(0,1)^{K\times n_p}$, where each row defines a soft-mask related to an anchor position to all the patch features, and each column records the weights of a feature to all the anchors. Moreover, we define multi-scale masks by adjusting $\delta$, 
	Finally, the hierarchical masks are represented by the collection $\mathbb{M}=\{\mathbf{M}(\delta_s)| s=1,2...,N\}$ with $N$ representing the number of scales, as shown in Fig. \ref{F.framework}g. The visualization of $\mathbb{M}$ can be found in the experiment section.
	
	\subsection{Kernel attention Transformer (KAT)} \label{S.kat}
\begin{figure}[t]
		\centering
		\includegraphics[width=0.9\textwidth]{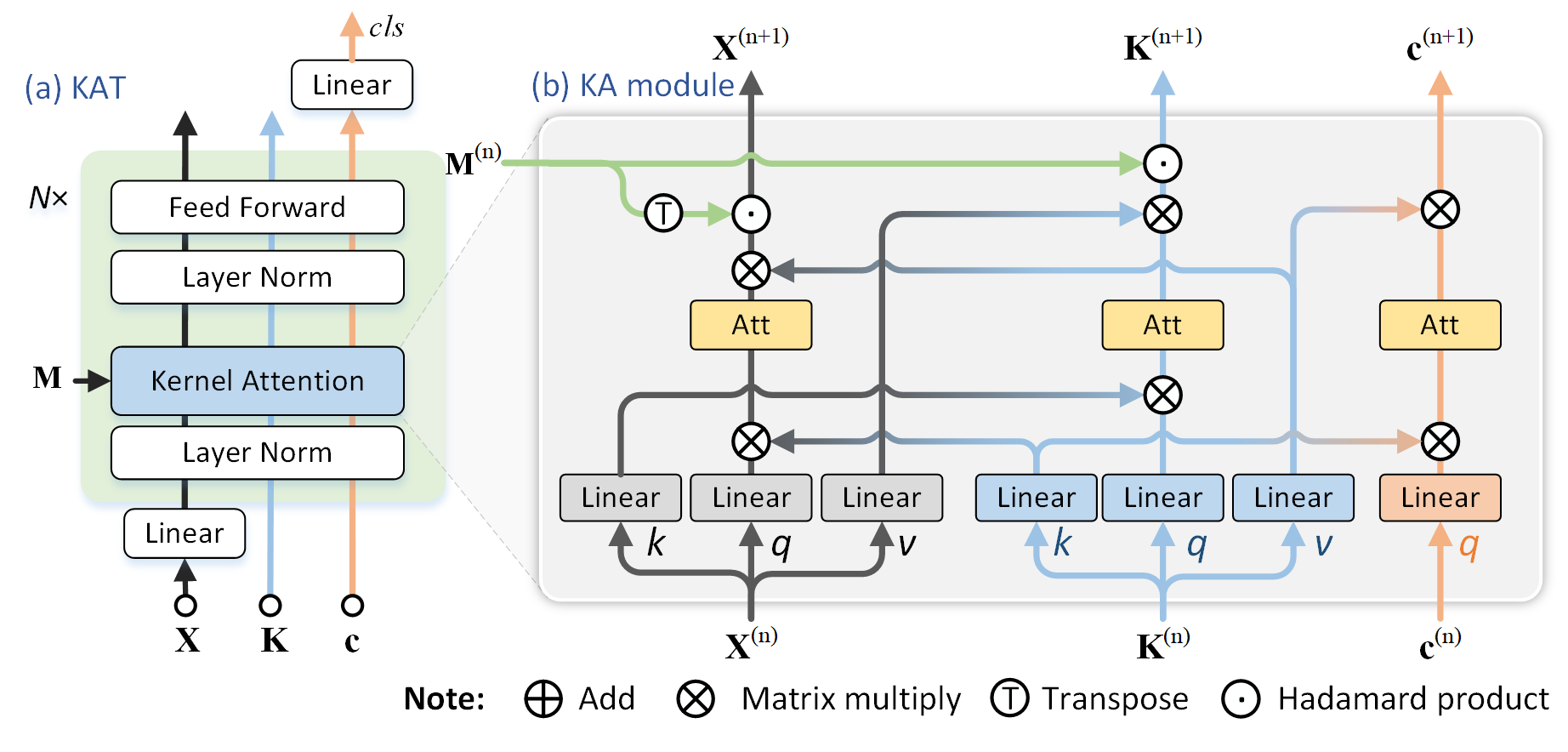}
		\caption{The structure of kernel attention network (KAT), where Feed Forward is composed of two linear projection layers, Att denotes a scaling operation followed by a softmax operation, $\mathbf{M}^{(n)}$, $\mathbf{X}^{(n)}$, $\mathbf{K}^{(n)}$, and $\mathbf{c}^{(n)}$ denotes the anchor masks, the representations of the patch tokens, kernel tokens, and classification token, respectively.} \label{F.network}
	\end{figure}

	The input of KAT includes the feature matrix $\mathbf{X}$ and the anchor masks $\mathbb{M}$ for each WSI. Each feature in $\mathbf{X}$ is regarded as a token. 
	As shown in Fig. \ref{F.network}a, the basic structure of KAT follows the ViT, which is constructed by stacking $N$ repeated blocks, which is composed of modules in the sequence of layer normalization, kernel attention, layer normalization, and feed-forward module, and the kernel attention and feed-forward module are with residual connections. Here, the kernel attention (KA) module is the core of KAT and is also the major difference of KAT to the common Transformer, which is detailed in this section.
	
	As shown in Fig \ref{F.network}b, the basic input of the $n$-th KA module includes the patch token representations $\mathbf{X}^{(n)}\in\mathbb{R}^{n_p\times d_e}$ and the classification token $\mathbf{c}^{(n)}\in\mathbb{R}^{d_e}$ with $d_e$ denoting the embedding dimension. Besides, we further define a set of tokens $\mathbf{K}^{(n)}\in\mathbb{R}^{K\times d_e}$ (as shown in Fig. \ref{F.framework}h) that serves as the kernels and correspondingly provides it the anchor masks $\mathbf{M}^{(n)}=\mathbf{M}(\delta_n)\in\mathbb{M}$. Instead of doing self-attention among the patch tokens, we propose to performing cross-attention between the kernels and the patch tokens. Namely, the information transmission in the KA module is achieved by a bi-direction message passing flow. One direction is the \textit{information summary flow (ISF)}, which is defined by equation
	\begin{equation}
	\mathbf{K}^{(n+1)} = \sigma\left(\mathbf{K}^{(n)}\mathbf{W}^{(n)}_{q}\cdot(\mathbf{X}^{(n)}\mathbf{W}^{(n)}_{k})^{\mathrm{T}}/\tau \right) \odot \mathbf{M}^{(n)}\cdot \mathbf{X}^{(n)}\mathbf{W}^{(n)}_{v},
	\end{equation}
	where the notations involving in $\mathbf{W}$ denote the trainable weights for linear projections, $\sigma$ denotes the softmax function, and $\tau$ is the scaling factor that is set the same as ViT.
	
	Another direction is the \textit{information distribution flow (IDF)}. The definition of \textit{IDF} is symmetric with \textit{ISF} by equation
	\begin{equation}
	\mathbf{X}^{(n+1)} = \sigma\left(\mathbf{X}^{(n)}\mathbf{W}^{(n)}_{q}\cdot(\mathbf{K}^{(n)}\mathbf{W}^{(n)}_{k})^{\mathrm{T}}/\tau\right) \odot \mathbf{M}^{\mathrm{T}(n)} \cdot \mathbf{K}^{(n)}\mathbf{W}^{(n)}_{v}.
	\end{equation}
	Through ISF, the individual representations of the patch tokens are reported to their nearby kernels for information summary. Then, through the IDF, the regional information carried by the kernels will be broadcast back to their nearby patches. Based on the bi-directional message passing flow, communication among the feature tokens of the WSI can be accomplished. 
	
	Correspondingly, in each KA module, we set aside a token to sum up the information from all the kernels for the purpose of classification. The message passing is defined as 
	\begin{equation}
	\mathbf{c}^{(n+1)} = \sigma\left(\mathbf{c}^{(n)}\mathbf{W}^{(n)}_{q}\cdot(\mathbf{K}^{(n)}\mathbf{W}^{(n)}_{k})^{\mathrm{T}}/\tau\right) \cdot\mathbf{K}^{(n)}\mathbf{W}^{(n)}_{v}.
	\end{equation}
	Supposing the number of the kernels is fixed, the computational complexity of KA module is $\mathcal{O}(n_p)$ to the token number $n_p$.
	
	At the end of the KAT, we built a linear layer on the output of the classification token for WSI classification. 
	The number of neurons of the last linear layer is the same as the type number of the WSIs. Finally, the entire KAT is trained end-to-end by cross-entropy loss function with Adam optimizer. The inputs of the kernel tokens and the classification token, i.e. $\mathbf{K}^{(0)}$ and $\mathbf{c}^{(0)}$, are randomly initialed and kept trainable in the training stage. To ensure all the kernels have consistent action for the same allocation of nearby features, we make all the kernels share the same set of trainable parameters. To further improve the performance of the KA module, we extended it to the Multi-head KA module following the paradigm of Transformer \cite{dosovitskiy2020image}.
	
	\section{Experiment and Result}
	The proposed method was evaluated on two large-scale WSI datasets. (1) \textit{Gastric-2K} contains 2040 WSIs of gastric histopathology from 2040 patients. These WSIs are categorized into 6 subtypes of gastric pathology, including Low/High-grade intraepithelial neoplasia, Adenocarcinoma, Mucinous adenocarcinoma, Signet-ring cell carcinoma, and Normal.
	(2) \textit{Endometrial-2K} contains 2650 WSIs of endometrium histopathology from 2650 patients. These WSIs are categorized into 5 subtypes of endometrial pathology, including Well/Moderately/Low-differentiated endometrioid adenocarcinoma, Serous endometrial intraepithelial carcinoma, and Normal.
	
	In each dataset, the WSIs were randomly separated into train, validation, and test subset by the proportion of 6:1:3. All the models discussed in the experiment were trained within the training set. The validation set was used to perform early stop and hyper-parameter tuning. And the results reported in the following sections were obtained on the test set by the final models.
	
	The window sliding and CNN feature extraction were performed on the WSI under $20\times$ lenses (the resolution is $0.48 \mu$m/pixel). The window size, as well as the image patch size, was set to $224\times224$ and the feature dimension $d_f=1280$ for the EfficientNet-b0 structure. 
	The proposed method was implemented in Python 3.8 with torch 1.9 and Cuda 10.2, and run on a computer with 2 Xeon 2.66GHz CPUs and 4 GPUs of Nvidia Geforce 3090. 
    More details please refer to the source code on \url{https://github.com/zhengyushan/kat}.
	\begin{figure}[t]
		\centering
		\includegraphics[width=0.8\textwidth]{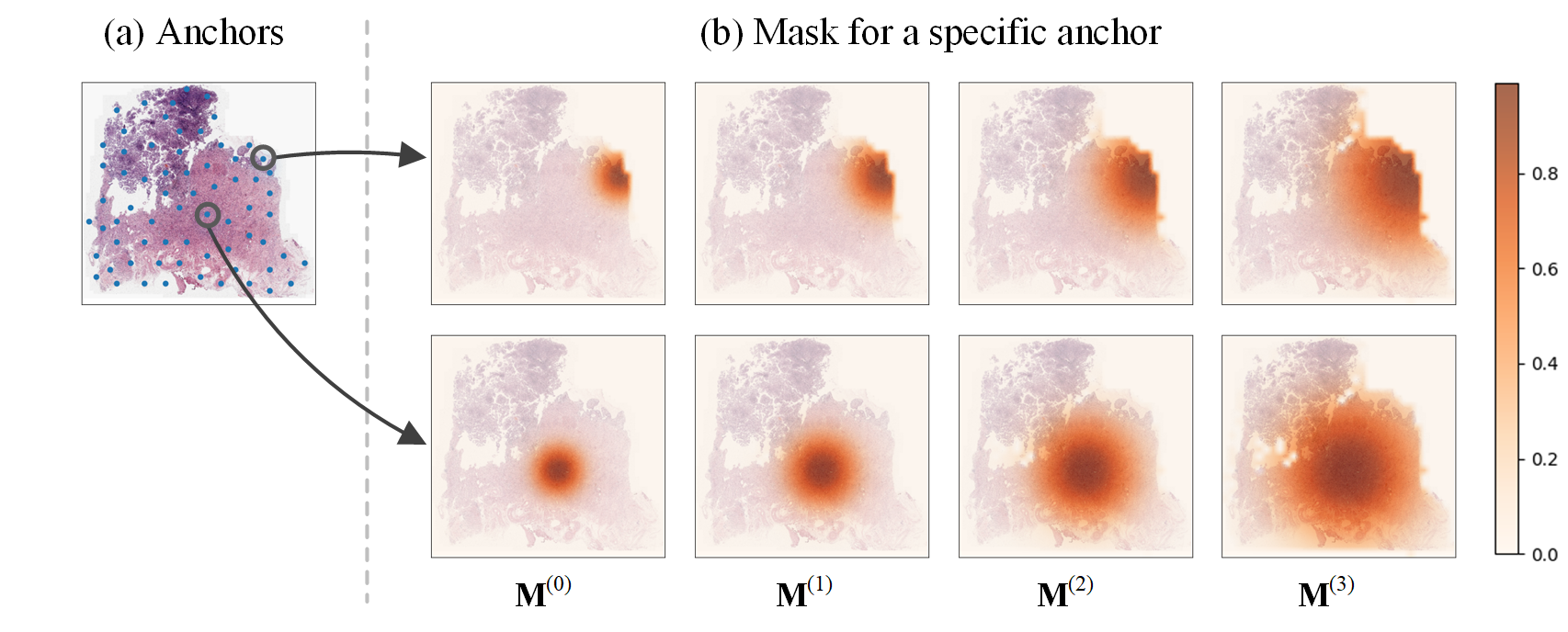}
		\caption{An instance of the anchor position allocation and the corresponding hierarchical masks under the setting $\bar{n}_k=144$ and $N=4$.} \label{F.mask}
	\end{figure}
	\subsection{Model verification}
	We first conducted experiments to verify the design of the proposed KAT model. The number of kernels $K$ is the main hyper-parameter that is additional to the common ViT. To make the proposed KAT be scalable to the size of WSI, we assigned an adaptive $K$ for each WSI by defining $K=[n_p/\bar{n}_k$], where $\bar{n}_k$ is the desired number of patches processed per kernel. Correspondingly, we empirically set $\delta_n^2=\bar{n}^2_k\cdot 2^n$. It makes the intersection of the (-$\delta_0$, +$\delta_0$) area of all the anchors basically cover all the patches, as shown in Fig. \ref{F.mask}. The performance of KAT with different $\bar{n}_k$ on the \textit{Endometrial-2K} dataset is presented on Table \ref{T.ablation}. The results show that KAT performs relatively consistent when $\bar{n}_k\in[64,256]$ and $\bar{n}_k=144$ is the most appreciate for the dataset.
	
	Then, we conducted the ablation study, for which the results are also summarized in Table \ref{T.ablation}. KAT w/o kernels denotes KAT without kernels and masks. In this situation, the KAT degrades to the common ViT structure. It causes a significant decrease in macro AUC for the subtyping from 0.835 to 0.788 and the AUC for the binary classification from 0.983 to 0.969. These results have demonstrated the effectiveness of the proposed kernel attention design and the corresponding anchor masking strategy. Furthermore, the FLOPs and GPU memory for the inference of a WSI respectively enlarge by 3.37$\times$ and 6.46$\times$. It shows the efficiency advantage of KAT over the normal ViT.
	KAT w/o masks indicates KAT without the $\mathbf{M}^{(n)}$, where the spatial information passing control for the patches and the anchors are discarded. The results show that it achieves a comparable performance to KAT w/o kernels, i.e., ViT, while maintaining a low computational complexity. In this case, KAT w/o masks acts as a linear approximation model for ViT like Linformer \cite{wang2020linformer}, Nystr\"{o}mformer \cite{xiong2021nystromformer}, etc. In the model KAT w/o IDF, we use self-attention operations among patch tokens to substitute the \textit{IDF} for patch token encoding. It causes a decrease of 0.012 to 0.013 in the AUCs. The results show that getting the information from the kernels is more efficient than that from individual patch tokens. It indicates the kernels contain higher-level semantics compared to the individual tokens.  
	
	\begin{table}[t]
		\caption{The results on the \textit{Endometrial-2K} dataset for different settings of KAT, where the classification accuracy (Acc.), and macro/weighted area under the receiver operating characteristic curve (mAUC/wAUC) for the sub-typing, and the specificity (Spec.), sensitivity (Sens.), and AUC for the binary task are calculated. The floating-point operations (\#FLOPs) and GPU memory cost (Mem.) per WSI are also provided.}\label{T.ablation}
		\centering
		\begin{tabular}{l|ccc|cccc|cc}
			\toprule[1pt]
			\multirow{2}*{Models} &\multicolumn{3}{c|}{\textbf{Sub-typing}} & \multicolumn{4}{c|}{\textbf{Normal vs. Others}}&\#FLOPs&Mem.\\
			&Acc.&mAUC&wAUC&Acc.&Spec.&Sens.&mAUC&($\times10^9$)&(MiB)\\
			\midrule[1pt]
			KAT ($\bar{n}_k=16$) &0.575 & 0.811 & 0.866 & 0.941 & \textbf{0.881} & 0.967 & 0.973 & 0.227& 899.71\\
			KAT ($\bar{n}_k=64$) &0.597 & 0.830 & 0.881 & 0.945 & 0.835 & 0.981 & 0.972 & 0.217 &581.56\\
			KAT ($\bar{n}_k=144$) & \textbf{0.608} & \textbf{0.835} & \textbf{0.882} & \textbf{0.949} & 0.867 & 0.975 & \textbf{0.983} & 0.213 & 569.31\\
			KAT ($\bar{n}_k=256$) & 0.582 & 0.825 & 0.872 & 0.945&0.834 & 0.981& 0.977 & \textbf{0.208} & 566.03\\
			\midrule[1pt]
			KAT w/o kernels & 0.549 & 0.788 & 0.840 & 0.925&0.810 & 0.963 & 0.969 & 0.701 & 3655.1\\
			KAT w/o masks & 0.567 & 0.814 & 0.857 &0.927 &0.822 & 0.963 & 0.960 & 0.213 & \textbf{563.24}\\
			KAT w/o IDF & 0.586 & 0.823 & 0.869 & 0.947 & 0.827 & \textbf{0.983} & 0.970 &0.743 & 3953.2\\
			\bottomrule[1pt]
		\end{tabular}
	\end{table}
	\subsection{Comparison with other methods}
\begin{figure}[t]
    	\centering
    	\begin{minipage}{0.49\linewidth}
    		\subfigure[Sub-typing task on Endometrial-2K]{\includegraphics[width=\textwidth]{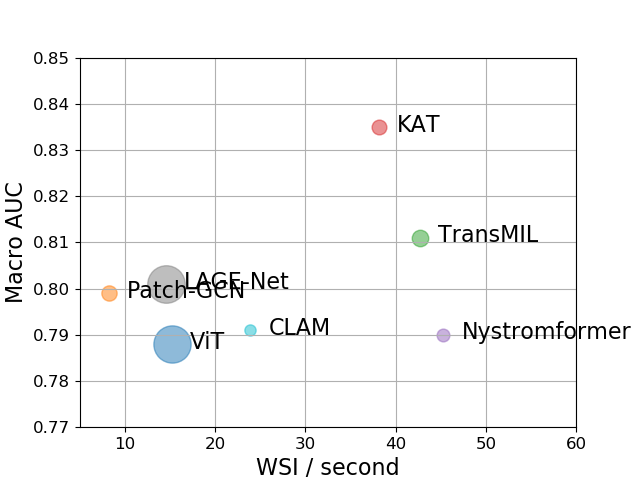}}
    	\end{minipage}
    	\begin{minipage}{0.49\linewidth}
    		\subfigure[Sub-typing task on Gastric-2K]{\includegraphics[width=\textwidth]{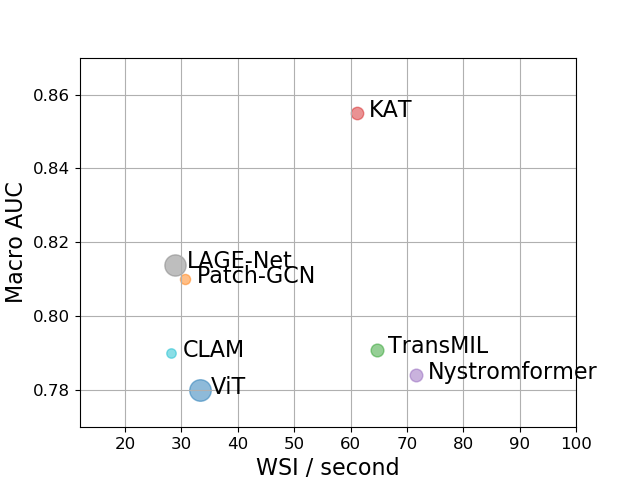}}
    	\end{minipage}
    	\begin{minipage}{0.49\linewidth}
    		\subfigure[Binary task on Endometrial-2K]{\includegraphics[width=\textwidth]{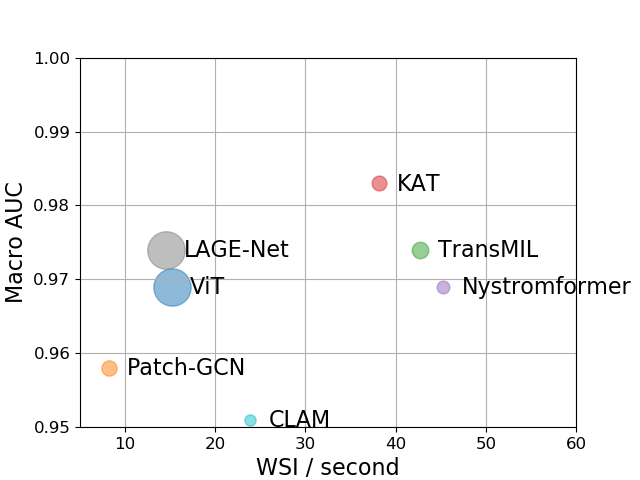}}
    	\end{minipage}
    	\begin{minipage}{0.49\linewidth}
    		\subfigure[Binary task on Gastric-2K]{\includegraphics[width=\textwidth]{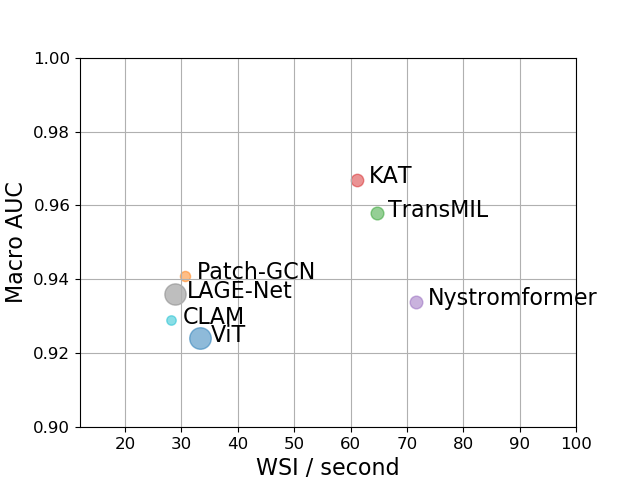}}
    	\end{minipage}
    	\caption{The joint plots of the macro AUC, the Speed of the inference, and the GPU memory cost per WSI on the two datasets, where the size of the dots is in direct proportion to the memory cost.} \label{F.result}
    \end{figure}

	The proposed method was compared with 6 methods \cite{chen2021whole,lu2021data,shao2021transmil,xiong2021nystromformer,dosovitskiy2020image,zheng2022encoding}. For all the Transformer-based models, we uniformly stacked four self-attention blocks with 8 heads and set the embedding dimension $d_e=256$. The results are presented in Table \ref{T.comparison} and Fig. \ref{F.result}
	
	\begin{table}[t]
		\scriptsize
		\caption{Comparison of the state-of-the-art methods, where the \textit{Speed} represents the number of WSI per second in the inference stage of the compared models.}\label{T.comparison}
		\centering
		\begin{tabular}{l|ccc|cccc|c}
			\toprule[1.5pt]
			\multirow{2}*{\textit{Endometrial-2K}} &\multicolumn{3}{c|}{\textbf{Sub-type classification}} & \multicolumn{4}{c|}{\textbf{Normal vs. Others}}&Speed \\
			&Accuracy&mAUC&mAUC&Accuracy&Specificity&Sensitivity&AUC&(WSI/s)\\
			\midrule[1pt]
			ViT \cite{dosovitskiy2020image} & 0.549 & 0.788 & 0.840 & 0.925&0.810 & 0.963 & 0.969 & 15.2\\
			Nystr\"{o}mformer \cite{xiong2021nystromformer} & 0.571 & 0.790 & 0.832 & 0.936 & 0.853 & 0.967 &0.969&\textbf{45.3}\\
			CLAM \cite{lu2021data} & 0.574 &0.791 &0.835 & 0.938 & 0.846 & 0.963 & 0.951&23.9 \\
			Patch-GCN \cite{chen2021whole} & 0.534 & 0.799 & 0.832 & 0.918 & 0.840 & 0.943 &0.958&8.1\\
			TransMIL \cite{shao2021transmil} &0.552&0.811&0.858&0.941&\textbf{0.855}&0.971&0.974&42.7\\
			LAGE-Net \cite{zheng2022encoding} & 0.598 & 0.801 & 0.855 & 0.938 &0.852 & 0.963 & 0.974&14.5\\
			KAT (Ours)& \textbf{0.608} & \textbf{0.835} & \textbf{0.882} &\textbf{ 0.949} & 0.830 & \textbf{0.985} & \textbf{0.983} &38.1\\
			\midrule[1.5pt]
			\multirow{2}*{\textit{Gastric-2K}} &\multicolumn{3}{c|}{\textbf{Sub-type classification}} & \multicolumn{4}{c|}{\textbf{(Normal+LGIN) vs. Others}}&Speed \\
			
			&Accuracy&mAUC&mAUC&Accuracy&Specificity&Sensitivity&AUC&(WSI/s)\\
			\midrule[1pt]
			ViT \cite{dosovitskiy2020image}& 0.765& 0.780 & 0.938 &0.842 & 0.827 & \textbf{0.913} & 0.924 & 33.3 \\
			Nystr\"{o}mformer \cite{xiong2021nystromformer}  & 0.819 & 0.784 & 0.935 & 0.900 & \textbf{0.953} & 0.760 &0.934& \textbf{71.7}\\
			CLAM \cite{lu2021data} & 0.788 & 0.790 &0.915 &  0.898 & 0.909 & 0.868  & 0.929&28.2\\
			Patch-GCN \cite{chen2021whole} & 0.797 & 0.810 & 0.939 & 0.874 & 0.915 & 0.770 & 0.941& 15.7\\
			TransMIL \cite{shao2021transmil} & \textbf{0.824} &0.791 & 0.944 & 0.902 & 0.940 & 0.796 & 0.958& 64.7\\
			LAGE-Net \cite{zheng2022encoding} &0.775 & 0.814 & 0.951 & 0.907 & 0.935 & 0.851 &0.936 & 28.9\\
			KAT (Ours)& 0.819 & \textbf{0.855} & \textbf{0.955} & \textbf{0.915} & 0.941 & 0.866 & \textbf{0.967}&61.2\\
			\bottomrule[1.5pt]
		\end{tabular}
	\end{table}
	
	Overall, the proposed method achieves the best performance with a mAUC / AUC of 0.835 / 0.983 in the sub-typing/binary classification task on the \textit{Endometrial-2K} dataset, which is 2.4\% / 0.9\% superior to the second-best methods, and a mAUC / AUC of 0.855 / 0.967 in the two tasks on the \textit{Gastric-2K} dataset, which is 4.1\% / 0.9\% superior to the second-best methods.
	
	Nystr\"{o}mformer \cite{xiong2021nystromformer} provides a linear solution to approximate standard self-attention in Transformer. The computational complexity is reduced to $\mathcal{O}(n)$. Meanwhile, the better convergence makes Nystr\"{o}mformer achieve higher classification accuracy than ViT \cite{dosovitskiy2020image}. Patch-GCN \cite{chen2021whole} is a typical spatial-graph-based WSI classification method, where the patches in the WSI are connected by a spatial graph. The adjacency information of the local patterns is considered by GCN in the encoding of the WSI. Therefore, it delivers comparable performance to ViT and Nystr\"{o}mformer. LAGE-Net \cite{zheng2022encoding} is a composition model of ViT and GCN, where both the long-range and adjacency relationship are considered.
	However, the combination usage of ViT and GCN causes an even lower computational efficiency than ViT. TransMIL \cite{shao2021transmil} utilizes the Nystr\"{o}mformer module to catch the long-range relationship in high computational efficiency. Meanwhile, a pyramid position encoding generator (PPEG) is built to extract the spatial relationship, which contributes to a better performance than Nystr\"{o}mformer. Nevertheless, the spatial relationship described by PPEG is inconsistent for different WSIs, especially for the gastric biopsy dataset where the tissue area varies a lot for different WSI. This made TransMIL be inferior to Patch-GCN and LAGE-Net on the gastric WSI sub-typing task. 
	
	In comparison, the proposed method builds uniformly distributed anchors that is adaptive to the shape and size of the tissue region and generates hierarchically region masks for the anchors to describe the local to the global relationship of the patches. The spatial relationship described by KAT is more complete and consistent compared to the previous methods and therefore achieves relatively better performance. Moreover, the kernel-based cross-attention computation in the KA module maintains a relatively high computational efficiency. The speed of inference is 38.1 WSIs per second on the endometrial dataset and 61.2 on the gastric dataset, which is comparable to TransMIL and Nystr\"{o}mformer.
	
	\section{Conclusion}
	In this paper, we have proposed a novel model named kernel attention Transformer (KAT) with the corresponding anchor mask generation approach for histopathology whole slide image classification. The experiments on two large-scale datasets have proven the KAT with the hierarchical anchor masks is both effective and efficient in the tasks of WSI sub-typing and binary classification, and achieves the state-of-the-art overall performance. Future work will develop representative region detection and encoding methods based on the kernel attention property of the KAT.
	
	\subsubsection*{Acknowledgments.}
	This work was partly supported by the National Natural Science Foundation of China [grant no. 61901018, 62171007, 61906058, and 61771031], partly supported by the Anhui Provincial Natural Science Foundation [grant no. 1908085MF210], and partly supported by the Fundamental Research Funds for the Central Universities of China [grant no. JZ2022HGTB0285].
	%
	%
	\bibliographystyle{splncs04}
	\bibliography{refs}

\end{document}